\documentclass[11pt,a4paper]{article}
\usepackage{authblk}
\usepackage[hyperref]{eacl2021}
\usepackage{times}
\usepackage{latexsym}
\usepackage{xcolor}
\usepackage{booktabs,siunitx}
\usepackage{textcomp}
\usepackage{graphicx}
\usepackage{hologo}
\usepackage{parskip}
\usepackage{array}
\usepackage{multirow}
\newcolumntype{C}[1]{>{\centering\arraybackslash}m{#1}}

\usepackage{microtype}
\usepackage{authblk}
\aclfinalcopy 

\title{Mono vs Multilingual BERT: A Case Study in Hindi and Marathi Named Entity Recognition}

\author{Onkar Litake$^{1,3}$, Maithili Sabane$^{1,3}$, Parth Patil$^{1,3}$, Aparna Ranade$^{1,3}$, \\\and 
Raviraj Joshi$^{2,3}$ \\

$^{1}$Pune Institute of Computer Technology, Pune\\
$^{2}$Indian Institute of Technology Madras, Chennai\\
$^{3}$ L3Cube, Pune\\

\texttt{onkarlitake@ieee.org,}
\texttt{\{msabane12,parthpatil8399,aparna.ar217\}@gmail.com}\\ 
\texttt{ravirajoshi@gmail.com}
}

\date{}

\begin{document}

\maketitle
\begin{abstract}
Named entity recognition (NER) is the process of recognising and classifying important information (entities) in text. Proper nouns, such as a person's name, an organization's name, or a location's name, are examples of entities. The NER is one of the important modules in applications like human resources, customer support, search engines, content classification, and academia. In this work, we consider NER for low-resource Indian languages like Hindi and Marathi. The transformer-based models have been widely used for NER tasks. We consider different variations of BERT like base-BERT, RoBERTa, and AlBERT and benchmark them on publicly available Hindi and Marathi NER datasets. We provide an exhaustive comparison of different monolingual and multilingual transformer-based models and establish simple baselines currently missing in the literature. We show that the monolingual MahaRoBERTa model performs the best for Marathi NER whereas the multilingual XLM-RoBERTa performs the best for Hindi NER. We also perform cross-language evaluation and present mixed observations.
\end{abstract}

\section{Introduction}

Named Entity Recognition \cite{nerfirst}, a term coined in 1995, refers to a popular technique of the information extraction process in natural language processing. It is a two-step process that involves (a) detection of a named entity and (b) categorization of the entity. These categories include a myriad of entities like names of persons, locations, organizations, numerical expressions like percentages, monetary values, and temporal values like date, time. The applications of these entity recognitions include text summarization \cite{textsum}, customer support \cite{customersupport}, machine translation \cite{translation}, efficient search algorithms \cite{searchalgo}, etc.

The process of NER can be performed in numerous ways. The English language has a large body of NER literature; however, very few efforts have been recorded in the Hindi and Marathi language. It is because of the deficit of well-annotated corpus and available tools. Additionally, these languages have various intricacies like the lack of capitalization and the equivocation between proper nouns and common nouns i.e., the Indian languages contain a myriad of common words which can be used as proper nouns. Furthermore, Hindi lacks a rigid grammar pattern. These shortcomings make it difficult to use the existing deep learning approaches that have successfully been used in English processing, to be used for these languages. Along with that, standard tools like Stanford NER \cite{stanfordner} and other POS and NER taggers do not have support for Hindi and Marathi. Recently, Hindi and Marathi text classification has received some attention \cite{joshi2019deep,kulkarni2022experimental,kulkarni2021l3cubemahasent,joshi2022l3cube,velankar2021hate}

Named Entity Recognition can mainly be performed using three major approaches. These include NER using Machine Learning \cite{nerml}, Rule-based NER \cite{nerrule}, and Hybrid NER \cite{nerhybrid}.  NER using Machine learning involves building a model using tagged text. Some examples include Conditional Random Fields (CRF) \cite{crf}, Support Vector Machine (SVM) \cite{svm}, Hidden Markov Model \cite{8004034} (HMM), and tools like Spacy \cite{spacy} and Stanford NER. Rule-based NER uses rules defined typically by linguists. It includes Lexicalised grammar, Gazetteer list, list of triggered words, and so on. Hybrid NER uses an amalgamation of machine learning and rule-based approaches. It could be a combination of HMM model with CRF or the Gazetteer method with HMM etc. 

Deep Learning \cite{deeplearning} is becoming increasingly popular as a result of its superior accuracy when trained with large amounts of data. Its architecture is also adaptable to new challenges. Deep Learning approaches outperform others when domain awareness is lacking for feature introspection since feature engineering is less of a concern and these techniques tend to solve problems end to end. Transformer-based systems have become increasingly popular in recent years due to their highly efficient architectures \cite{transformer,inproceedings1}. In this paper, we try to establish baseline numbers on various publicly available datasets for Marathi and Hindi languages by training various transformer architectures. We make use of pre-trained BERT-based masked language models. The multilingual variants of these language models have been very popular recently for low-resource languages. We also try to provide a comparative study of multilingual and monolingual variants of these language models. The monolingual variants are only pre-trained on Hindi or Marathi data. We are focusing on transformer architectures like BERT\cite{bert}, RoBERTa \cite{roberta}, and their variants such as mBERT, RoBERTa-Hindi, Indic Bert \cite{indicbert}, mahaBERT \cite{joshi2022l3cube} etc. We also perform a cross-language evaluation of these BERT models since both Hindi and Marathi share the Devanagri script.
The main contribution of this work is as follows:
\begin{itemize}
    \item We show that monolingual Marathi models based on mahaBERT perform better than their multilingual counterpart thus showing the importance of language-specific pre-training.
    \item For the Hindi language the multilingual models perform better hence there is a need to develop better monolingual language models.
    \item During cross-language evaluation the results favor Marathi models. The Marathi monolingual models based on the mahaBERT scale well on Hindi NER datasets but the same is not true for publicly available Hindi BERT models as they perform poorly on Marathi NER datasets. Again highlights the need to have better resources for Hindi.
\end{itemize}
The rest of the paper is structured as follows. Section 2 surveys the advancement in Named Entity Recognition and focuses on Indian languages. Section 3 explains how we set up our experiments to test various models. Section 4 summarizes the findings from all of the experiments. Our paper's conclusion is presented in Section 5.

\section{Related Work}

The concept of Named Entity Recognition originated in 1995 at the Message Understanding Conferences (MUC)\cite{nerfirst} in the US. However, it was not until 2008 that the study on Indian languages received widespread recognition. 

\begin{figure}[h!]
    \frame{\includegraphics [scale= 0.27]{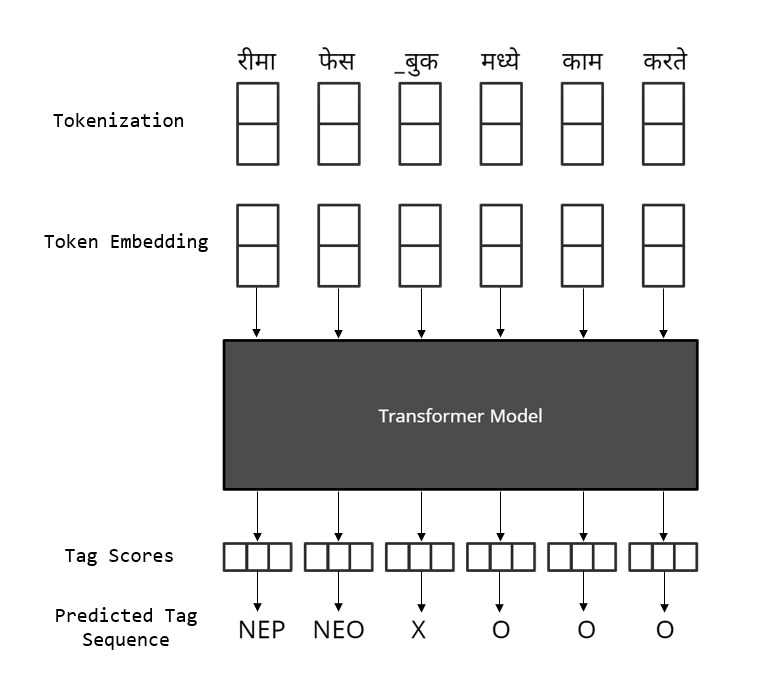}}
  \centering\caption{Model Architecture}
  \label{fig:1}
\end{figure}

Krishnarao et al.\cite{krishnarao2009comparative} presented a comparative study based on two algorithms viz. Support Vector Machine (SVM) and Conditional Random Field(CRF). The paper provides a comparison between these algorithms that are trained using the same data. The CRF model is found to be superior to the SVM model. 
Srihari et al.\cite{srihari} used a hybrid system consisting of a combination of handmade rules, the Hidden Markov Model, and MaxEnt. Such hybrid systems were found to be more effective while performing NER.
\\\\ 
Subsequently, as the advancement continued, deep learning models were used to perform the NER task. The most popularly used NER models were Convolutional Neural Network (CNN)\cite{cnn}, Long-Short Time Memory(LSTM)\cite{lstm}, Bi-directional Long-Short Time Memory(BiLSTM)\cite{bilstm}, Transformers.

Shah\cite{shah} emphasizes the NER techniques used so far for various Indian languages. The paper provides a comparative analysis of the methods used for identifying named entities. It compares the Hidden Markov Model (HMM) method and the Conditional Random Field(CRF) method and finds that the CRF method is the most effective approach for Indian languages. 
Bhattacharjee et al.\cite{bhatta} examine various techniques for NER in Indian languages, with a focus on Hindi. It compares the Machine Learning (ML), Rule-based, and Hybrid techniques. The paper aims to identify the gaps in the existing NER systems, especially in the Hindi language, as these systems are trained to perform on predetermined datasets and do not produce results on universal datasets. The paper determines that the machine learning approach is more systematic while predicting entities that are not known, but it showcases an accuracy lower than the rule-based system. 
Patil\cite{patil} discusses the importance of NE recognition for Marathi along with the concerns and obstacles that come with NE recognition in the Marathi language. It also looks at various methodologies and techniques for creating learning resources that are necessary for extracting NEs from unstructured natural language material.

Further, in the deep learning domain, a variety of models have been proposed and tested to perform the NER tasks on Indic languages including SVM by Singh et al.\cite{singh}, Conditional Random Field by Shishtla et al.\cite{shishtla}, and Hidden Markov model. But with the advancement of deep learning architectures, methods were proposed to identify entities from text without adhering to language-specific rules. 
However, Shah et al.\cite{shah} and Shelke et al.\cite{Shelke2020ANA} have illustrated encouraging results by utilizing BiLSTM networks to simplify the NER complexities. Our work builds upon theirs and adds other models that are trained to obtain the required accuracies. Additionally, we have performed the NER task on the Marathi dataset.

Murthy et al.\cite{judi} showcases the influence of the differences in tag distributions of common named entities between primary and helping languages on the efficacy of multilingual learning. The paper proposes a measure based on symmetric KL divergence using neural networks like CNNs and Bi-LSTMs to filter out the highly divergent training examples in the helping language to solve this challenge. 

\section{Experimental Setup}

\subsection{Dataset}
We are limiting our work of performing Named Entity Relation(NER) tasks on Hindi and Marathi which are among the top 3 languages spoken in India. We are carrying out NER tasks on 
all publicly available datasets for these languages.\cite{inproceedings}

For Hindi, we are using datasets released in IJCNLP\cite{ijcnlp} in 2008 and  WikiAnn NER Corpus released by Pan et al.\cite{wiki_bio} in 2017. The IJCNLP dataset contains a total of 11,400 sentences. It has a total of 12 categories named as a person, organization, location, abbreviation, brand, title-person, title-object, time, number, measure, designation, terms. No split for the data was provided, hence we split the data into 70-15-15 train, test, tune respectively. We have corrected a few tags which were improperly annotated. For example, The ‘Term’ entity is ununiformly tagged throughout the dataset. It is tagged using 'B-NETE', 'I-NETE', 'B- nete', 'B-N ETE', 'I-NETE/=', and many more such tags. We have replaced all of them with either ‘B-NETE’, or ‘I-NETE’ to keep the tagging uniform. We have also discarded ambiguous tags and replaced them with ‘O’ tags. Following is the list of such tags: 'B- NET','B-NEB', 'B-NET/=','B- NET','B-NEB','B-NET/=', 'B-NTA','B-Terms','B-k1','I- NET', 'I-NEB', 'I-NET/=', 'I-k1'. The dataset released by Pan et al. for Hindi contains a total of 11,833 sentences and has been divided into 3 categories namely Organization, Person and Location. It is a "silver-standard" dataset.

For Marathi, we are using a dataset released by Murthy et al.\cite{judi} named IIT Bombay Marathi NER Corpus in 2018 and  WikiAnn NER Corpus released by Pan et al.\cite{wiki_bio} in 2017. The dataset contains a total of 5,591 sentences. It has a total of 3 categories named Location, Person, Organization. Train-Test-Tune split of the dataset was provided beforehand. Both the datasets used were in IOB format. The dataset released by Pan et al. for Marathi contains a total of 14,978 sentences and has been divided into 3 categories namely Organization, Person and Location. It is a "silver-standard" dataset.

We removed the IOB formatting from the IJCNLP 200 NER Corpus in accordance with the previous work carried out on it. Challenges faced while working with these datasets were: 
\begin{itemize}
    \item The IJCNLP dataset and IIT Bombay Marathi NER Corpus included English words.
    \item Tagging was non-uniform in the IJCNLP dataset and also included some ambiguous tags.
    \item More than  68 percent of the sentences in the IJCNLP dataset and 39 percent of sentences in the IIT Bombay Marathi NER Corpus included sentences that had only O tags
\end{itemize}

There are more datasets for Hindi and Marathi on NER but they aren’t publicly available. Following are the names of such datasets. a) FIRE-2013- Named-Entity Recognition Indian Languages. b) FIRE 2014 - Named-Entity Recognition Indian Languages. c) FIRE 2015- Entity Extraction from Social Media Text Indian Languages (ESM-IL) d)FIRE 2016 - Shared Task on Code Mix Entity Extraction in Indian Languages (CMEE-IL) e) TDIL- Named Entity Annotated Corpora for Marathi. f) TDIL- Named Entity Corpora for Hindi, Marathi, Punjabi.

\textbf{Example:}
\begin{figure}[h!]
    \includegraphics [scale= 0.27]{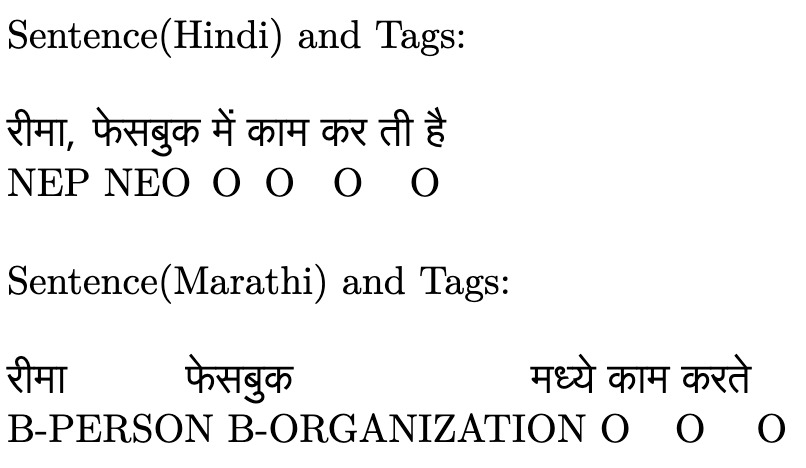}
\end{figure}

\begin{table*}
\centering
\scalebox{0.8}{
  \begin{tabular}{|l|l|l|l|l|l|l|}
    \hline
    \multirow{2}{*}{Dataset} &
      \multicolumn{3}{c}{\quad Count of Sentences } &
      \multicolumn{3}{c|}{\quad Count of Tags}\\
      
    & \quad Train \quad\quad & \quad Test \quad\quad  &\quad  Tune  \quad\quad&\quad  Train \quad\quad & \quad Test \quad\quad &\quad  Tune\quad\quad  \\
    \hline
    
    IJCNLP 200 NER Corpus &\quad 7979 & \quad 1711 & \quad 1710 &\quad  208750 & \quad 44692\quad  &\quad 44146 \\
    \hline
   IIT Bombay Marathi NER Corpus \qquad& \quad 3588 & \quad 1533 & \quad 470 &\quad  67775 & \quad 32214 &\quad 8370 \\
    \hline
    WikiAnn NER Corpus(Marathi)  &\quad  10674 &\quad  4304 & \quad -- & \quad 76006 & \quad 32572 &\quad\quad --  \\
    \hline
    WikiAnn NER Corpus(Hindi) & \quad 8356 &\quad  3477 &\quad-- &\quad 48601 &\quad  20829 &\quad\quad --\\
    \hline
  \end{tabular}}
  \centering\caption{Count Of sentences and tags in the datasets}
\label{table:1}
\end{table*}

\begin{table*}
\centering
\scalebox{0.8}{
  \begin{tabular}{ | m{2cm} | m{2cm}| m{2cm} | m{2cm}| } 
    \hline
 
 Tags & Train & Test & Tune \\ 
 \hline
  O & 18172 & 38692 & 38049\\
  NETE & 6468 &1315 & 1379\\
  NEN & 4529 & 1062 & 1009\\
  NEP  &3893 & 886 & 806\\
  NEL &3257 & 692 & 880\\
  NEO &2119 & 459 & 392\\
  NETI &2210 & 482 & 524\\
  NEM &2017 & 463 & 550 \\
  NETO &1630 & 397 & 348\\
  NED &796 & 141 & 145 \\
  NEA &459 & 103 & 64\\
 \hline
  \end{tabular}}
  \centering\caption{Count of individual tags of IJCNLP 200 NER Corpus }
\label{table:2}
\end{table*}

\begin{table*}
\centering
\scalebox{0.8}{
\begin{tabular}{ | m{2cm} | m{2cm}| m{2cm} | m{2cm}| } 
 \hline
 
 Tags & Train & Test  \\ 
 \hline
  O & 20015 &  9243 \\
  I-ORG &  8195 & 4638 \\
  B-ORG &2907 & 1512 \\
  I-PER &  7445 & 1705 \\
  B-PER  &5570 & 973 \\
  I-LOC &2314 & 1224 \\
  B-LOC & 2155 &1534 \\
  
 \hline
\end{tabular}}
\centering\caption{Count of individual tags of WikiAnn NER Corpus (Hindi) }
\label{table:3}
\end{table*}

\begin{table*}
\centering
\scalebox{0.8}{
  \begin{tabular}{ | m{5cm} | m{2cm}| m{2cm} | m{2cm}| } 
 \hline
 
 Tags & Train & Test & Tune \\ 
 \hline
  O & 61235 & 28215 & 7349\\
  B-LOCATION & 3372 & 1871 & 598\\
  I-LOCATION & 1449 & 1277  &198\\
  B-PERSON & 974 &432 & 131\\
  I-PERSON & 572 & 328 & 63\\
  I-ORGANIZATION & 98  & 56 & 16\\
  B-ORGANIZATION &75 &35 &15\\
 \hline
\end{tabular}}
\centering\caption{Count of individual tags of IIT Bombay Marathi NER Corpus}
\label{table:4}
\end{table*}

\begin{table*}
\centering
\scalebox{0.8}{
\begin{tabular}{ | m{2cm} | m{2cm}| m{2cm} | m{2cm}| } 
 \hline
 
 Tags & Train & Test  \\ 
 \hline
  O & 46011 &  19633 \\
  I-ORG &   7076 & 3047 \\
  B-ORG &3053 &   1263 \\
  I-PER &  6686 & 3020 \\
  B-PER  &4469 & 1651 \\
  I-LOC &3019 & 1330 \\
  B-LOC &  5692 & 2628\\
  
 \hline
\end{tabular}}
\centering\caption{Count of individual tags of WikiAnn NER Corpus (Marathi) }
\label{table:5}
\end{table*}

\subsection{Model Architecture}
In natural language processing, the Transformer seeks to solve sequence-to-sequence tasks while also resolving long-range dependencies. The Transformer NLP model includes an "attention" mechanism that analyzes the association between all the words in a sentence. It generates differential weightings to suggest which components in the sentence are most important for determining how a word should be interpreted. This accounts for the quick and efficient resolution of ambiguous elements. For example, the input given is a sentence, the transformer recognizes the context that grants the meaning of each word in the sentence. As the feature improves parallelization, the training time is reduced. The general model setup is shown in Figure \ref{fig:1}.
\\
	\textbf{\\BERT}: Originating from the pre-training contextual representations, BERT is a transformer-based technique for NLP pre-training developed by Google. It is a deep bidirectional model, meaning that it grasps the details from both sides of a token's context while training. The most important characteristic of BERT is that it can be fine-tuned by adding a few output layers. 
\\
\textbf{\\mBERT}: The next stage in developing models that grasp the meaning of words in context is MBERT, which stands for multilingual BERT. By simultaneously encoding  all of their information on MBERT a deep learning model was trained on 104 languages. 
\\
\textbf{\\ALBERT}: Google AI open-sourced ALBERT, a transformer architecture based on BERT which uses much fewer parameters than the state-of-the-art model BERT model. As compared to BERT models, these models have higher data throughput and can train about 1.7 times faster than the BERT model. IndicBERT, a multilingual ALBERT model, trained on large-scale datasets covers 12 major Indian languages. Many public models like mBERT and XLM-R contain more parameters as compared to IndicBERT, yet the latter performs very well on a variety of tasks.
\\
\textbf{\\RoBERTa}: RoBERTa is a self-supervised transformers model, trained on a large corpus of English data. This implies that it was pre-trained on raw texts solely, with no human labeling, and then used an automatic method to build labels and inputs from those texts.
XLM-RoBERTa is a multilingual model that has been trained in 100 languages. It does not require lang tensors to recognize which language is used, unlike some XLM multilingual models. It is also capable of determining the proper language from the input ids.

\section{Results}

\begin{table*}
\centering
\scalebox{0.8}{
\begin{tabular}{lSSSSSSSS}
    \toprule
    \multirow{2}{*}{Dataset} &
      \multicolumn{4}{c}{IIT Bombay} &
      \multicolumn{4}{c}{WikiAnn} \\
      & {F1} & {Precision} & {Recall} & {Accuracy} & {F1} & {Precision} & {Recall} &{Accuracy} \\
      \toprule
    Multicase BERT & 58.35 & 63.67 & 54.58 & 92.42 & 86.49 & 86.25 & 86.73 & 95.18 \\
    \hline
    Indic BERT & 60.79 & 66.05 & 53.76 & 92.57 & 87.03 & 87.06 & 87.00 & 95.13 \\
    \hline
    Xlm-Roberta & 62.32 & 64.14 & 60.60 & 93.00 & 87.38 & 86.92 & 87.85 & 95.48 \\
    \hline
    Roberta-Marathi & 43.81 & 42.64 & 45.03 & 91.34 & 82.00 & 80.26 & 83.82 & 93.73 \\
    \hline
    MahaBERT &  62.57 & 64.67 & 60.61 & 92.97 & 88.18 & 88.22 & 88.14 & 95.77 \\
    \hline
    MahaRoBERTa & \textbf{64.34} & 65.64 & 63.08 & 92.90 & \textbf{88.90} & 88.59 & 89.20 & 96.06 \\
    \hline
    MahaAlBERT & 60.00 & 63.77 & 56.52 & 92.52 & 87.15 & 87.19 & 87.11 & 95.14 \\
    \hline
    RoBERTa Hindi & 42.19 &41.52  &42.88  &91.10  &82.50  &81.69  &83.33  & 95.29\\
    \hline
    Indic-transformers\\-hi-roberta &36.80  &36.81  &36.7  &90.49  &80.00  &78.73  &81.32  &94.27 \\
    \bottomrule
 \end{tabular}}
  \centering\caption{F1 score(macro), precision and recall of various transformer models using the Marathi datasets.}
\label{table:6}
\end{table*}

\begin{table*}
\centering
\scalebox{0.8}{
\begin{tabular}{lSSSSSSSS}
    \toprule
    \multirow{2}{*}{Dataset} &
      \multicolumn{4}{c}{IJCNLP 200 } &
      \multicolumn{4}{c}{WikiAnn} \\
      & {F1} & {Precision} & {Recall} & {Accuracy} & {F1} & {Precision} & {Recall} &{Accuracy} \\
      \toprule
    Multicase BERT & 72.74 & 70.64 & 74.97 & 95.09 & 81.21 & 79.64 & 82.85 & 91.50 \\
    \hline
    Indic BERT & 71.63 & 70.50 & 72.87 & 95.12 & 82.65 & 81.44 & 83.90 & 92.01 \\
    \hline
    Xlm-Roberta & \textbf{75.90} & 74.19 & 77.70 & 95.62 & \textbf{83.04} & 82.80 & 83.27 & 91.68 \\
    \hline
    Roberta-Hindi & 69.06 & 67.27 & 70.95 & 95.29 & 80.52 & 78.52 & 82.63 & 90.85 \\
    \hline
    indic-transformers\\-hi-roberta & 64.36 & 62.44 & 66.41 & 94.27 & 73.79 & 70.71 & 77.15 & 87.75\\
    \hline
    Roberta-Marathi  & 61.69 & 59.39 & 64.17 & 94.48 & 79.46 & 77.20 & 81.85 & 90.82 \\
    \hline
    MahaBERT & 72.91 & 70.71 & 75.21 & 95.09 & 81.95 & 80.99 & 82.93 & 91.71 \\
    \hline
    MahaRoBERTa & 75.30 & 73.64 & 77.04 & 95.61 & 80.66 & 79.71 & 81.63 & 91.37 \\
    \hline
    MahaAlBERT & 69.83 &69.62  &70.03  &94.82  &81.68  &80.87  &82.50  &92.23  \\
    \bottomrule
  \end{tabular}}
   \centering\caption{F1 score(macro), precision and recall of various transformer models using the Hindi datasets.}
\label{table:7}
\end{table*}
 In this section, we report and discuss the F1 score obtained by training various models 
 \footnote{Multicase BERT: https://huggingface.co/bert-base-multilingual-cased \\
Indic BERT: https://huggingface.co/ai4bharat/indic-bert \\
Xlm-roberta: https://huggingface.co/xlm-roberta-base \\
Roberta-Marathi: https://huggingface.co/flax-community/roberta-base-mr \\
Roberta-Hindi: https://huggingface.co/flax-community/roberta-hindi \\
Indic-transformers-hi-roberta: https://huggingface.co/neuralspace-reverie/indic-transformers-hi-roberta \\
MahaBERT: https://huggingface.co/l3cube-pune/marathi-bert \\
MahaRoBERTa: https://huggingface.co/l3cube-pune/marathi-roberta \\
MahaAlBERT: https://huggingface.co/l3cube-pune/marathi-albert-v2}
 on respective datasets. Table 6 represents the results of transformer models trained on the IIT Bombay Marathi NER Corpus and WikiAnn NER Corpus (Marathi). Table 7 represents the results of IJCNLP 200 NER Hindi Corpus and  WikiAnn NER Corpus(Hindi).
 
 The number of sentences in the training set for Hindi is double that of Marathi, due to which the models trained on the Hindi dataset have a better F1 score compared to that of Marathi.  MahaRoBERTa model which has RoBERTa as its base architecture performs the best on both the Marathi datasets. Roberta-Marathi in spite of being trained in Marathi has the least F1 score. The monolingual Marathi models based on MahaBERT perform better than the multi-lingual models. Whereas for Hindi NER datasets the multilingual models perform better than the Hindi monolingual counterparts. During the cross-language evaluation, we test the Marathi models on Hindi datasets and vice versa. This is desirable as both Marathi and Hindi share the same Devanagari script. We obeserved that models based on mahaBERT perform competitively on Hindi datasets. However, this is not true of Hindi models as they perform poorly on Marathi datasets. Both the models released by flax-community do not perform well on either of the languages. We, therefore, highlight the need for developing better resources for the Hindi language. In general, we observe that language-specific fine-tuning does not necessarily guarantee better performance and the factors underlying this disparity need to be investigated.
One hypothesis for this disparity could be attributed to the fact that the Marathi models have been trained on top of multi-lingual models whereas Hindi models have been trained from scratch.

\section{Conclusion}

Many NER systems have been deployed in English and other major languages. However, there hasn't been much work done on Hindi and Marathi languages. This study seeks to examine transformer-based deep learning-based NER solutions to Hindi and Marathi NER tasks. We benchmark for a host of monolingual and multi-lingual transformer-based models for Named Entity Recognition that includes multilingual BERT, Indic BERT, Xlm-Roberta, mahaBERT, and others. We show that monolingual training doesn't necessarily ensure superior performance. Although Marathi monolingual models perform the best same is not true for Hindi. Moreover, we observe that the mahaBERT models even generalize well on Hindi NER datasets. It is worthwhile to investigate the poor performance of mono-lingual models and is left to future scope.  

\section*{Acknowledgements}
This work was done under the L3Cube Pune mentorship program. We would like to express our gratitude towards our mentors at L3Cube for their continuous support and encouragement.

\bibliography{main}
\bibliographystyle{acl_natbib}




\end{document}